%% file: sample-sigconf-authordraft.tex
\documentclass[sigconf]{acmart}

\usepackage{ragged2e} 
\usepackage{booktabs,makecell, multirow, tabularx}
\usepackage{xspace}
\usepackage{xcolor}
\usepackage{enumitem}
\usepackage{graphicx}
\usepackage{subcaption}
\usepackage{amsthm}
\usepackage[ruled, lined, linesnumbered, commentsnumbered, longend]{algorithm2e}
\usepackage[inkscapelatex=false]{svg}
\usepackage{subcaption} 
\usepackage{float}
\usepackage{cleveref}

\newtheorem{problem}{Problem}

\newcommand{\modelname}{GuARD\xspace}
\newcommand{\modelnamesem}{GuARD+sem\xspace}
\newcommand{\modelnamegraph}{GuARD+graph\xspace}

\newcommand{\component}{Modality}
\newcommand{\rankonename}{ChatGLM-IND\xspace}
\newcommand{\vpara}[1]{\vspace{0.07in}\noindent\textbf{#1 }}

\newcommand{\hide}[1]{}

\newcommand{\cb}[1]{\textbf{\color{red}[(cb: #1 )]}}  

\AtBeginDocument{%
  }

\setcopyright{acmlicensed}

\copyrightyear{2025}
\acmYear{2025}
\setcopyright{cc}
\setcctype{by}
\acmConference[KDD '25]{Proceedings of the 31st ACM SIGKDD Conference on Knowledge Discovery and Data Mining V.2}{August 3--7, 2025}{Toronto, ON, Canada}
\acmBooktitle{Proceedings of the 31st ACM SIGKDD Conference on Knowledge Discovery and Data Mining V.2 (KDD '25), August 3--7, 2025, Toronto, ON, Canada}
\acmDOI{10.1145/3711896.3736993}
\acmISBN{979-8-4007-1454-2/2025/08}

\begin{document}

\title{\modelname: Effective Anomaly Detection through a Text-Rich and Graph-Informed Language Model}

\author{Yunhe Pang}
\authornote{Equal contribution.}
\authornote{Work was done when Yunhe interned at Zhipu AI.}
\affiliation{%
  \department{School of Computer Science and Engineering}
  \institution{Sun Yat-Sen University}
  \city{Guangzhou}
  \country{China}}
\email{pangyh8@mail2.sysu.edu.cn}


\author{Bo Chen}
\authornotemark[1]
\affiliation{%
  \department{Department of Computer Science and Technology}
  \institution{Tsinghua University}
  \city{Beijing}
  \country{China}}
\email{cb21@mails.tsinghua.edu.cn}

\author{Fanjin Zhang}
\authornotemark[1]
\authornote{Fanjin Zhang and Jie Tang are the corresponding authors.}
\affiliation{%
  \department{Department of Computer Science and Technology}
  \institution{Tsinghua University}
  \city{Beijing}
  \country{China}}
\email{fanjinz@tsinghua.edu.cn}

\author{Yanghui Rao}
\affiliation{%
  \department{School of Computer Science and Engineering}
  \institution{Sun Yat-Sen University}
  \city{Guangzhou}
  \country{China}}
\email{raoyangh@mail.sysu.edu.cn}

\author{Evgeny Kharlamov}
\affiliation{%
  \department{Bosch Center for Artificial Intelligence}
  \institution{Robert Bosch GmbH}
  \city{Renningen}
  \country{Germany}}
\email{Evgeny.Kharlamov@de.bosch.com}

\author{Jie Tang}
\authornotemark[3]
\affiliation{%
  \department{Department of Computer Science and Technology}
  \institution{Tsinghua University}
  \city{Beijing}
  \country{China}}
\email{jietang@tsinghua.edu.cn}
\renewcommand{\shortauthors}{Trovato et al.}

\begin{abstract}
Anomaly detection on text-rich graphs is widely prevalent in real life, such as detecting incorrectly assigned academic papers to authors and detecting bots in social networks.
The remarkable capabilities of large language models (LLMs) pave a new revenue by utilizing rich-text information for effective anomaly detection.
However, simply introducing rich texts into LLMs can obscure essential detection cues and introduce high fine-tuning costs.
Moreover, LLMs often overlook the intrinsic structural bias of graphs which is vital for distinguishing normal from abnormal node patterns.
To this end, this paper introduces \modelname, a text-rich and graph-informed language model that combines key structural features from graph-based methods with fine-grained semantic attributes extracted via small language models for effective anomaly detection on text-rich graphs.
\modelname is optimized with the progressive multi-modal multi-turn instruction tuning framework in the task-guided instruction tuning regime tailed to incorporate both rich-text and structural modalities. 
Extensive experiments on four datasets reveal that \modelname outperforms graph-based and LLM-based anomaly detection methods,
while offering up to $5\times$ speedup in training and $10\times$ speedup in inference
over vanilla long-context LLMs
on the large-scale WhoIsWho dataset.
\end{abstract}



\maketitle

\newcommand\kddavailabilityurl{https://github.com/THUDM/WhoIsWho/tree/main/mind}
\ifdefempty{\kddavailabilityurl}{}{
\begingroup\small\noindent\raggedright\textbf{KDD Availability Link:}\\
The source code of this paper has been made publicly available at \url{\kddavailabilityurl}.
\endgroup
}
\input{intro}

\input{relatedwork}

\input{problem}

\input{method}

\input{experiment}

\input{conclusion}

\begin{acks}
This work has been supported by the NSFC for Distinguished Young Scholar (62425601)  and New Cornerstone Science Foundation through the XPLORER PRIZE.
Yanghui Rao was supported by the National Natural Science Foundation of China (62372483).
This work is also supported by the Natural Science Foundation of China (NSFC) 62406164, the Postdoctoral Fellowship Program of CPSF under Grant Number GZB20240358 and 2024M761680.
\end{acks}

\bibliographystyle{ACM-Reference-Format}
\bibliography{sample-base}

\appendix


\input{appendix}

\end{document}

%% file: intro.tex
\section{Introduction}
\label{sec:intro}

\begin{figure}[t]
  \begin{subfigure}[b]{\columnwidth}
    \centering
    \includegraphics[ width=\columnwidth]{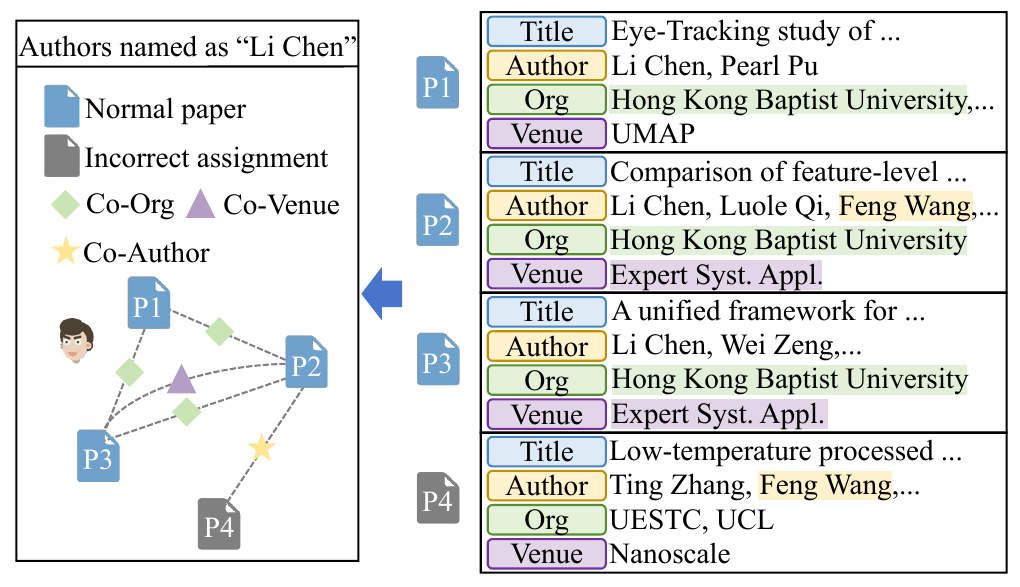}
    \caption{Anomaly detection on the academic network for detecting incorrect assignment papers to authors. } 
    \label{fig:IND}
  \end{subfigure}
    \begin{subfigure}[b]{0.40\columnwidth}
    \centering
    \includegraphics[width=\linewidth]{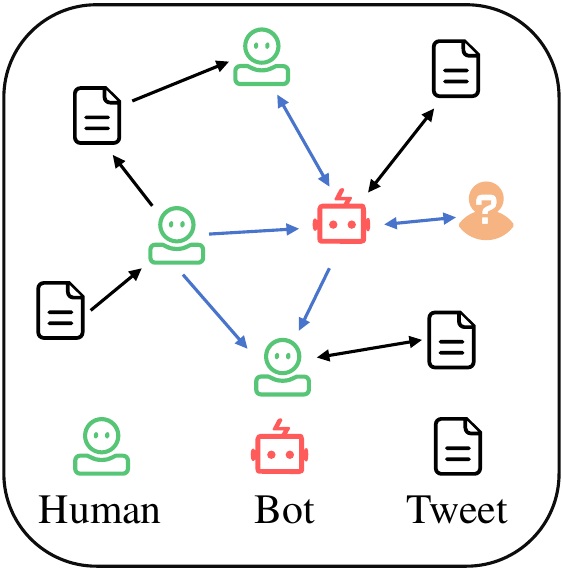}
    \caption{Bot Detection}
    \label{fig:sub-second}
  \end{subfigure}
  \hfill  
  \begin{subfigure}[b]{0.58\columnwidth}
    \centering
    \includegraphics[width=\linewidth]{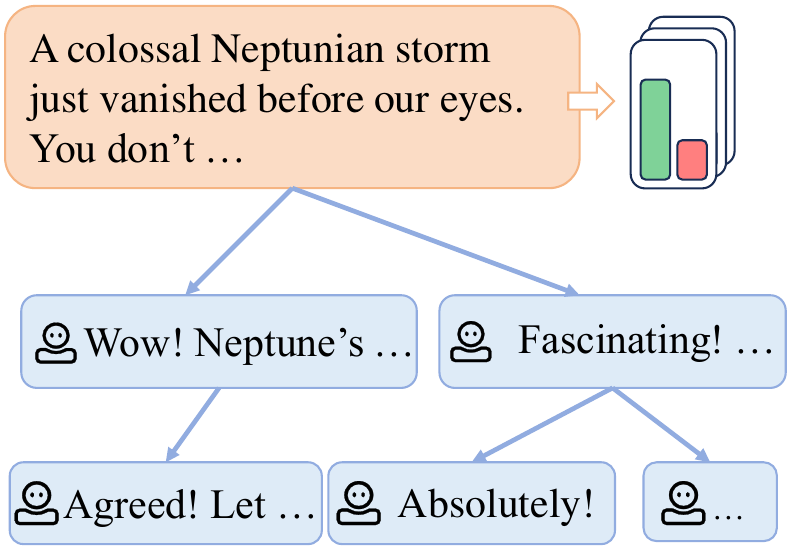}
    \caption{Misinfomation Detection}
    \label{fig:sub-third}
  \end{subfigure}
  \caption{Illustration about typical anomaly detection in text-rich graph: \textnormal{(a) shows the anomaly detection in authors' multi-relation graph (i.e. Incorrect Assignment Detection, IND); (b) showcases the bot detection on social networks; (c) depicts the misinformation detection of news articles on comment networks.}}
  \label{fig:problem}
\end{figure}

Anomaly detection in text-rich graphs arises in many real-world applications, ranging identifying incorrectly assigned academic papers
to authors with ambiguous names~\cite{whoiswho,zhang2019oag},
to the detection of bots and misinformation in social networks~\cite{DBLP:conf/iclr/ChenS24,
DBLP:conf/emnlp/LucasUYLR023}.
These scenarios are becoming increasingly common due to the substantial proliferation of research papers and the influx of abundant AI-generated content on the web.
Taking academic networks~\cite{tang2008arnetminer,zhang2019oag,zhang2022oag} as an example (Figure \ref{fig:problem}(a)), detecting incorrect assignment papers in the academic network requires not only a nuanced understanding of the semantic coherence between papers and authors, but also the structural roles emphasized by graph topology. Other applications like bot detection and misinformation detection (Figure \ref{fig:problem}(b,c)) face similar challenges.

Existing endeavors commonly focus on either detecting outliers purely based on the structural traits or discerning abnormal signals from informative text spans~\cite{DBLP:conf/icml/TangLGL22,DBLP:conf/nips/TangHGZL23,adllm}, which pays less attention to anomaly detection on real-world text-rich graphs.
Among these attempts, GCCAD~\cite{gccad} contrasts each node with the global context to learn discriminative node embeddings.
LMBot~\cite{DBLP:conf/wsdm/CaiT0ZWZL24} iteratively distills knowledge from graph neural network(GNN) to small language model (SLM) to allow GNN and SLM to mutually enhance each other.
However, these graph-based methods focus more on key structural features and implicitly incorporate short
textual features via input node features, but lack characterizing the fine-grained semantic features embodied in node attributes.



Recently, large language models (LLMs)~\cite{brown2020language,llm} have shown remarkable performance across a wide range of natural language understanding and generation tasks, underpinned by their ability to capture fine-grained correlations through self-attention.
Several efforts~\cite{zhang2024oag,rank1,rank3} employ LLMs as the backbone and design specialized instruction templates for anomaly detection tasks. 
Despite these advances, LLM-based methods may struggle when confronted with massive input texts. For example, in the realm of 
detecting incorrect paper assignments within academic networks,
a productive author can have over $1\small{,}000$ papers and each paper possesses rich attributes (e.g., title, venue, author list, etc.).
For fake news detection, some news articles can be thousands of words long.
Processing such long context input can dilute critical detection signals and greatly increase fine-tuning costs.
Moreover, LLMs typically do not capture intrinsic structural biases in graph topologies.
Therefore, a holistic model that can effectively harness both the long-text attributes and structural relations in text-rich graphs is in great demand.
\begin{figure}[t]
    \centering
    \includegraphics[width=0.9\linewidth]{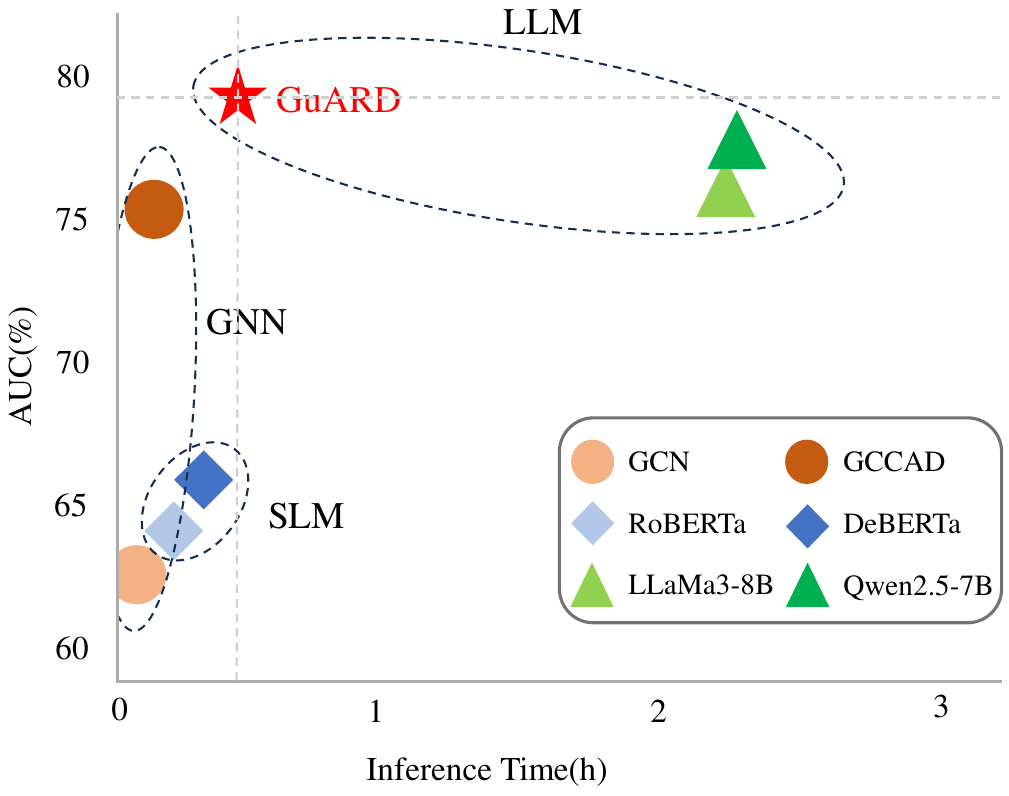}
    \caption{AUC vs. inference time of different models on the WhoIsWho dataset. GNN\textnormal{: Graph Neural Network, }SLM\textnormal{: Small Language Model, }LLM\textnormal{: Large Language Model.}
    }
    \label{fig:model}
\end{figure}

\noindent \textbf{Present Work.} 
Inspired by the aforementioned insights, we propose \modelname, a 
text-rich and graph-informed
language model for anomaly detection that combines the strengths of 
key structural feature extraction endowed by graph-based methods
and the fine-grained semantic features characterized by language models via effective multi-modal fusion.
Practically, we adopt a multi-modal-like multi-turn instruction tuning framework to incorporate each modality step by step as follows:
\begin{enumerate}[leftmargin=*]
    \item \textbf{Task-Guided Multi-Turn Instruction Tuning:}
    In the first stage, we aim to align the language model backbone to tackle the anomaly detection task via an instruction-tuning regime. 
    Specifically, we design a task-specific instruction template, as shown in Figure~\ref{fig:model}, 
    which takes the optional global context and target nodes as input and asks the LLMs
    to generate the label token supervised by ground truths.
    To foster context sharing and improve training/inference efficiency,
    we further design a multi-turn chat instruction template,
    which significantly improves the accuracy and efficiency of our model.
    
    \item \textbf{Semantic Embedding Module with Rich Attributes:}
    Limited by the context length, LLMs can only take nodes' key features as input. 
    To incorporate more informative nodal textual attributes into our framework,
    we employ a semantic embedding module that extracts and summarizes rich textual attributes through a small pre-trained language model,
    and then utilize a text projector to obtain a special text token that serves as the summarized semantic feature of each node.
    
    \item \textbf{Structural Embedding Module:}
    To endow LLMs with the capacity to capture structural information,
    we employ a structural embedding module that
    extracts and summarizes structural features from the graph-based methods,
    and then utilizes a graph projector to obtain a special graph token that acts as the structural feature of each node.
\end{enumerate}

Through these three successive stages of 
multi-modal-like instruction tuning,
both structural features and rich semantic characteristics are dynamically and effectively fused for 
robust anomaly detection.
Extensive experimental results on four datasets highlight the superiority of our proposed method.
Compared to advanced fine-tuned LLMs, 
\modelname achieves better or comparable anomaly detection accuracy, 
while significantly improving fine-tuning and inference time efficiency ($1.15\times$-$5\times$ speedup in training and  $3\times$-$10\times$ speedup in inference w.r.t. LoRA fine-tuned LLMs).

%% file: relatedwork.tex
\section{Related Work}
\label{sec:relatedwork}
Anomaly detection includes a variety of forms, including time-series anomaly detection, graph anomaly detection, and text/log anomaly detection. 
This work primarily focuses on anomaly detection in scenarios with rich-text data and underlying graph structures.
\subsection{LM-based Anomaly Detection}
\label{sec:lm_anomaly_detection}
Language model-based approaches for anomaly detection leverage the semantic distributional differences in textual data to identify anomalous attributes or outlying instances. These methods have been successfully applied in tasks such as spam detection, fraud detection~\cite{abkenar2023learning,wu2018twitter}, bot detection~\cite{feng2021twibot}, and log analysis~\cite{guan2024logllm}. 
ChatGLM-IND~\cite{zhang2024oag} defines an author's paper list as the global context.
It trains LLMs to detect anomalies by assessing the similarity between the global and local contexts to determine whether a paper is anomalously attributed to an author.
LMBot~\cite{DBLP:conf/wsdm/CaiT0ZWZL24} enhances the capabilities of language models by iteratively fine-tuning a combination of language models and graph neural networks. This joint optimization improves the model's ability to differentiate between human and bot-generated text effectively.
BotSay~\cite{botsay} adopts an in-context learning (ICL) paradigm with LLMs. By incorporating information from a user's ego-networks,
the model becomes better informed and guided in the task of bot detection.
DELL~\cite{DELL} employs a language model-based framework to enhance textual data via relevance analysis. It utilizes the outputs of a large language model to augment the original text and then fine-tunes the DeBERTa~\cite{deberta} model for text representation learning.

\subsection{Graph-based Anomaly Detection}
\label{sec:graph_anomaly_detection}
Graph anomaly detection~\cite{gccad,DBLP:journals/tkde/MaWXYZSXA23,DBLP:conf/cikm/DouL0DPY20}, compared to text-based anomaly detection, places greater emphasis on identifying structural anomalies within graph data.
Graph anomaly detection has been widely applied in scenarios with a large number of nodes and complex connections, such as social networks and citation networks.
GCCAD~\cite{gccad} introduces a contrastive learning framework for nodes and edges, which enhances network representation learning by identifying and removing incorrectly connected edges. This approach has demonstrated effectiveness in tasks such as financial fraud detection and identifying anomalous paper assignments.
In the task of bot detection in social networks, node classification models based on heterogeneous graph neural networks, such as HGT and RGCN~\cite{DBLP:conf/www/HuDWS20,rgcn,simple_hgn}, have shown strong performance by effectively leveraging the heterogeneity and structural dependencies in the graph.
Moreover, certain text-based anomaly detection methods construct graph structures via data augmentation techniques, either by modeling relationships between samples or within samples. These graph-based representations are often employed to provide additional signals for detecting anomalies in textual sequences.
In addition, some studies~\cite{DELL} construct comment networks and employ structural representation learning methods to incorporate non-textual features into the misinformation detection task. 
Leveraging graph structures alongside textual content improves the model's effectiveness in identifying misinformation.

Despite extensive studies on anomaly detection,
existing graph-based and small pre-trained language model-based methods have been found to lag behind LLM-based approaches~\cite{zhang2024oag}.
However, recent LLM-based methods struggle to efficiently and effectively utilize long-text information and graph structure information.
In this paper, we propose \modelname, a text-rich and graph-informed language model that effectively combines key textual features, rich textual attributes, and structural features in a progressive manner.

%% file: problem.tex
\section{Problem Definition}
\label{sec:problem}

In this section, we introduce 
the problem formulation of 
anomaly detection in text-rich graphs.

    



\begin{definition}
    \textbf{Text-Rich Graphs.}
    A text-rich graph is a graph $G = (\mathcal{V}, \mathcal{E}, \mathcal{S})$,
    where each node $v_i \in \mathcal{V}$ is associated with multiple attributes that make up a long text sequence $s_i \in \mathcal{S}$ 
    and $\mathcal{E}$ represents the set of edges between nodes.
\end{definition}

\begin{problem}
    \textbf{Anomaly Detection on Text-Rich Graphs.}
    Given a text-rich graph $G = (\mathcal{V}, \mathcal{E}, \mathcal{S}, \mathcal{Y})$,
    where $\mathcal{Y}$ is the set of node labels with 
    $y_i \in \mathcal{Y}$ be equal to $1$ if node $v_i$ is abnormal and $0$ otherwise,
    the goal is to find a function $f: \mathcal{V} \to \{0, 1\}$
    to determine whether a given node is an anomaly or not.
\end{problem}

\hide{
\begin{definition}
    \textbf{Paper.} A paper \( p \) is associated with multiple attributes, i.e., \( p = \{x_1, \ldots, x_F\} \), in which \( F \) denotes the number of attributes (e.g., title, venue, authors).
\end{definition}

\begin{definition}
    \textbf{Author.} An author \( a \) contains a paper set, i.e., \( a = \{p_1, \ldots, p_n\} \), where \( n \) is the number of papers authored by \( a \).
\end{definition}

\begin{problem}
    \textbf{Incorrect Assignment Detection (IND)}~\cite{whoiswho}. Given a conflated author entity \( {a^* = \{ p_i, \ldots, p_j, p_a, \ldots, p_b, \ldots, p_m, \ldots, p_n\}} \) coming from multiple authors \( \{a_1, \ldots, a_k\} \),
    where \( a_1 = \{p_i, \ldots, p_j\} \), \( a_2 = \{p_a, \ldots, p_b\} \), \( a_K = \{p_m, \ldots, p_n\} \),
    assuming \( a_1 \) covers the highest percentage of papers within \( a^* \), 
    the objective is to detect outlier papers owned by $\{a_2, ..., a_k\}$.
\end{problem}
}

%% file: method.tex

\begin{figure*}[t]
    \centering
    \includegraphics[width=16cm]{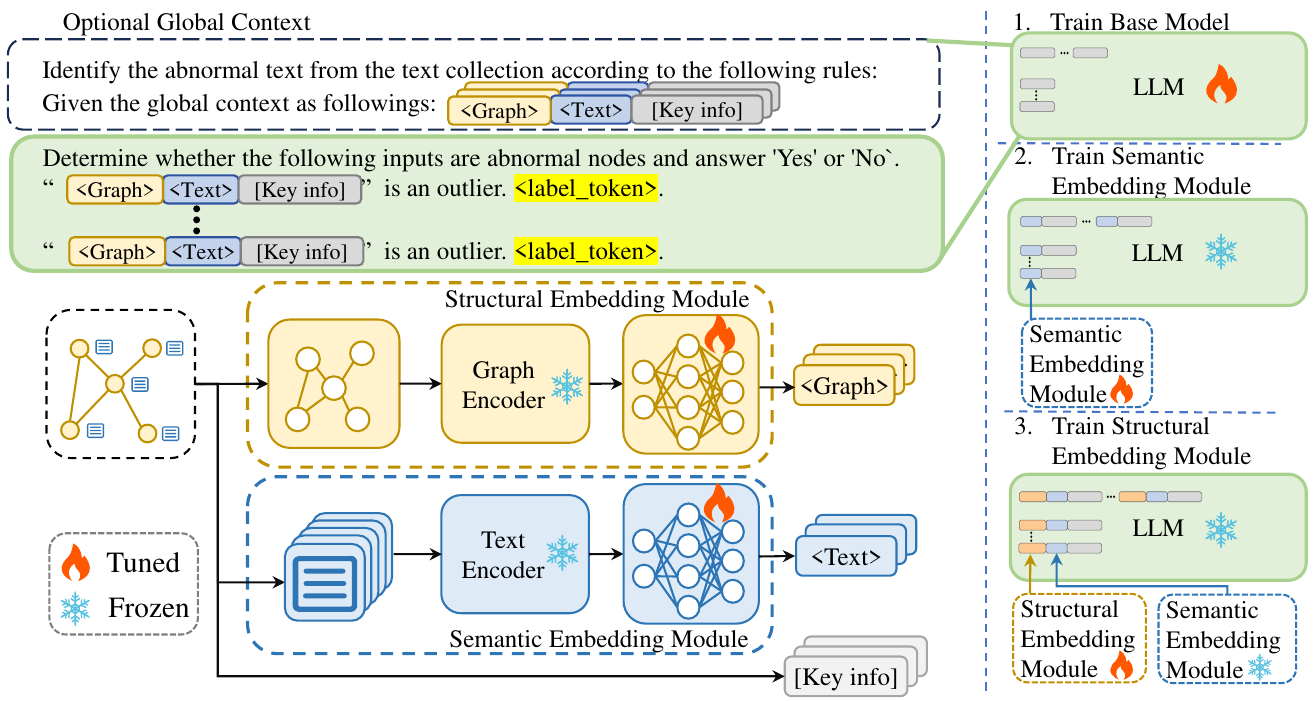}
    \caption{The entire workflow of \modelname in detecting anomalies on text-rich graphs. \textnormal{The entire instruction template comprises three parts of input: 
    1) key information (e.g. title) from each node (e.g. paper) is used as direct input tokens of the LLM;
    2) textual embedding provided by the semantic embedding module serves as a summarized embedding of the rich attributes of each node by replacing the token \texttt{<text>};
    3) structural embedding from the structural embedding module is incorporated into the LLM by replacing the token \texttt{<graph>} in LLM input. 
    Our framework trains different modules progressively in each modality to predict a special token \texttt{<label\_token>}.
    }}
    \label{fig:model}
\end{figure*}

\section{Model Framework}
\label{sec:Model architecture}
Previous anomaly detection methods on text-rich graphs typically adopt pre-trained language model-based methods or graph-based methods.
Graph-based methods fall short of capturing the semantic coherence across text-rich nodes.
Although large language models~\cite{roberta,meta2024introducing}, which formalize the anomaly detection problem as a question-answering task, 
excel in capturing intricate semantic correlations between nodes, they usually cannot extract crucial detection information among the massive input text effectively (e.g. $> 10$K). Furthermore, the intrinsic capacity of LLMs struggles to capture structural patterns of graphs.

In view these challenges, we propose \modelname, a text-rich and graph-informed language model to detect anomalies on text-rich graphs.
\textbf{Firstly}, \modelname introduces an effective LLM-based multi-turn instruction tuning framework,
which takes the key textual information of multiple nodes as input, enabling the basic capacity of detecting anomalies (see Section \ref{subsec:base}).
\textbf{After that}, we enhance this by introducing other rich attributes into the LLM through a semantic embedding module. This module summarizes the rich text attributes of each node, projecting them as semantic tokens into the LLM (see Section \ref{subsec:textemb}).
\textbf{Subsequently}, we employ a graph embedding module to integrate structural features, converting them into structural tokens for ingestion by the LLM (see Section \ref{subsec:graphemb}).
\textbf{Lastly}, we adopt a progressive training algorithm to ensure these modalities---key textual, semantic, and graph-based---are effectively combined (see Section \ref{Triple-Stage}).
In the following sections, we delve into the details of each component.

\subsection{Task-Guided Multi-Turn Instruction Tuning}
\label{subsec:base}
Prior arts~\cite{gccad,zhang2024oag} usually adopt graph-based methods for the anomaly detection problem.
However, recent attempts~\cite{zhang2024oag,adllm} leverage the LLMs as the backbone, holding the promise that it can capture semantic coherence among rich attributes to distinguish the anomaly patterns, which demonstrate the strong capacity of detecting anomalies. 
Inspired by this, we formalize the anomaly detection problem as a question-answering task, define an instruction template by 
incorporating the node attributes and its global contexts (if any), and then ask the LLM to answer whether the given node is an anomaly or not.
Noted that, the global context of the text-rich graph is not always necessary. In certain scenarios, the anomaly can be distinguished by contrast it with the surrounding global context~\cite{gccad}.
Taking the academic network as an example, a paper is detected as incorrectly assigned when it has a topic discrepant from the assigned author's profile, while for social bot detection,
each node does not possess a global context.

To convert the anomaly detection task into question-answering instructions, a straightforward idea is to feed all textual attributes of each node and then ask whether this node is an anomaly or not.
However, this solution raises the following concerns.
On the one hand, the long-context LLMs still struggle to identify relevant information and face the ``lost in the middle'' issue~\cite{DBLP:journals/tacl/LiuLHPBPL24}.
On the other hand, fine-tuning and inference of long-context LLMs incur significant time costs and memory costs.

To this end, we choose key textual attributes of each node as the LLM input. For different scenarios, where the textual information varies, we also select different information as the key attributes (described in section \ref{sec:details}).
If a node’s anomaly depends on its global context, we randomly sample a fixed number of nodes in the global context from the graph and prepend their key attributes to the instruction template. We then append the target node’s attributes, prompting the LLM to determine the anomaly via a custom token, \texttt{<label\_token>}.
However, due to the inefficiency of predicting anomalies one node at a time, 
we design a multi-turn instruction template that utilizes the shared contextual information across different nodes to be detected. 
Figure \ref{fig:model} illustrates the corresponding instruction template.
By stacking multiple local queries into the input, we generate multiple predictive results for multiple nodes in a single auto-regressive decoding step.
This significantly reduces the model training and inference time, with only a minor computational overhead from limited key contextual inputs.
The training objective is defined as: 
\begin{equation}\mathcal{L} = -\sum_{i=1}^{N} \log p(w_i \mid \text{context}_i)\label{eq:loss}\end{equation}
where \(N\) denotes the number of stacking target nodes 
and \(w_i\) denotes the logits of the ground-truth label of \texttt{<label\_token>}.
Note that LLMs typically use a causal masking format in their self-attention, allowing subsequent queries to incorporate information from earlier ones and thus providing richer context.
Consequently, the detection information of earlier queries can serve as few-shot examples for the prediction of label tokens in latter queries. This mechanism leverages both the correlation between the target node and its global context and the relevance of preceding queries, enhancing the prediction accuracy for latter queries.

During the inference phase, we directly take the normalized logit of token ``Yes'' and token ``No'' as the final logit of \texttt{<label\_token>}. 
The final logit is calculated by:
\begin{equation}y_i=\frac{z_i^y}{z_i^y + z_i^n}\end{equation}
where \(z_i^y\) and \(z_i^n\) denote the logit of ``Yes'' and ``No'' for the $i$-th query, respectively.

\subsection{Semantic Embedding Module} 
\label{subsec:textemb}

Due to the context-length limitation of LLMs, it is infeasible to incorporate all rich textual attributes of each node into LLMs. Principally, longer context lengths demand substantially increased GPU memory and computational cost.
As as result, only the key features can be selected as the basic detection information into LLMs as detailed in Section~\ref{subsec:base}.
To comprehensively leverage the remaining rich textual information, we propose a semantic embedding module.
Specifically, it adopts a small pre-trained language model to convert textual attributes into a sequence of embeddings. 
Additionally, an adaptive pooling module is used to summarize these embeddings, and a text projector is employed to align the feature space of the semantic embedding module with that of LLMs.

For each input node,
a pre-trained language model (PLM)~\cite{devlin2018bert,roberta} is leveraged as a text encoder to derive token-level representations of these nodes. 
Specifically, 
let \(X_i\) denote the 
$i$-th
node in the graph, 
and \(X_i=\{{x_i^0,x_i^1,...,x_i^n}\}\), 
where \(x_i^j\) denotes the $j$-th token in the text sequence.  
We obtain the textual embedding by applying mean pooling to the PLM-embedded text sequence and further use a two-layer feed-forward network (FFN) with a Swish activation function~\cite{swish,silu} as the projector to align the dimension of the PLM's token representation with the input dimension of the LLM.
The entire process of the semantic embedding module and the alignment can be represented as follows:
\begin{align}
\{h_i^0,h_i^1,...,h_i^n\} = \text{PLM}&(\{x_i^0,x_i^1,...,x_i^n\})
\label{eq:plm}\\
H_i=\text{mean-pooling}&(\{h_i^0,h_i^1,...,h_i^n\})
\label{eq:mean pooling} \\
H_i'=\text{FFN}_{\text{Swish}}(H_i)\label{eq:text proj}
\end{align}
\noindent where $h_i^j$ denotes the hidden representation of token $x_i^j$
and $H_i'$ is the summarized semantic embedding.

\hide{
We obtain the encoded hidden token representation of the text sequence:
\begin{equation}\{h_i^0,h_i^1,...,h_i^n\} = 
\text{PLM}
(\{x_i^0,x_i^1,...,x_i^n\}).\label{eq:plm}\end{equation}

\noindent where $n$ is the input token length
and $h_i^j$ denotes the hidden representation of token $x_i^j$.
Subsequently, a mean pooling function is employed to summarize the representation of the paper:
\begin{equation}H_i=
\text{mean-pooling}
(\{h_i^0,h_i^1,...,h_i^n\}).
\label{eq:mean pooling}
\end{equation}

To align the representation space of the PLM to that of the LLM,
an additional projector module is required. 
For simplicity and effectiveness,
a two-layer feed-forward network with a Swish activation function~\cite{silu} is employed as the text projector:
\begin{equation}H_i'=\text{FFN}_{\text{Swish}}^T(H_i).
\label{eq:text proj}
\end{equation}
}

To integrate the summarized semantic features into the LLM input,
an external token \texttt{<text>} is introduced and strategically positioned at the beginning of each node's input text.
By replacing the original embedding of this token in the LLM with the embedding $H_i'$
projected from the semantic embedding module,
the supplementary textual features are effectively incorporated into the LLM input.

\subsection{Structural Embedding Module}

\label{subsec:graphemb}

Existing attempts ~\cite{gccad,zhang2024oag} demonstrate that
graph neural networks (GNNs) excel at detecting anomalies through a comprehensive understanding of the node features and graph topology.
However, the existing LLMs are not adept at capturing structural information.
To address this gap, we train a state-of-the-art GNN-based anomaly detection method such as GCCAD~\cite{gccad} or HGT~\cite{DBLP:conf/www/HuDWS20}
to obtain the structural-enhanced node and graph embeddings.
Then the node and graph representations are concatenated to form the structural features of each node. 
Finally, a graph projector is utilized to align the structural features with the hidden space of LLMs. 

Let $\mathcal{A}$ denote the adjacency matrix of the input  graph
and $X_{\text{nd}}$ denote the node input features. 
The procedure of the structural embedding module can be formalized as follows:
\begin{align}
Z^g,\{Z_0,Z_1,...,Z_m\}&=
\text{GNN}
(\mathcal{A}, X_{\text{nd}}) 
\label{eq:graph encoder} \\
Z_i'= \text{FFN}_{\text{Swish}}& (\text{concat}(Z_i,Z^g))\label{eq:graph projector}\end{align}
where \(Z^g\) and \(Z_i\) denotes the learned graph embedding and the $i$-th node's hidden embedding, 
respectively. Then the concatenated node and graph embedding are fed into a 2-layer FFN projector to obtain the input graph embedding for the LLM. 
Correspondingly, we define a similar special graph token \texttt{<graph>}, 
which is positioned prior to \texttt{<text>} and added into the input of LLM.
The original embedding of \texttt{<graph>} is replaced by the embedding $Z_i'$.

\subsection{Progressive Instruction-tuning}
\hide{
\cb{Introducing the algorithm execution snippet here would be better.}
}
\label{Triple-Stage}
To effectively intergrate the three modalities features---key textual, semantic, and graph---into our framework, 
we employ a three-phase progressive instruction-tuning procedure.
In the first phase as detailed in Section~\ref{subsec:base},
we begin with training the base LLM utilizing parameter-efficient fine-tuning~\cite{lora,li2021prefix,peft}, endowing it with basic anomaly detection capabilities.

In the second phase, we freeze the parameters of the previously trained LLM and further introduce the semantic embedding module. 
As the pre-trained language model can yield meaningful semantic features through the 1st-phase fine-tuning, we fix its parameter here and focus on optimizing the text projector at this stage.

In the third phase, we freeze all parameters from the preceding stages and concentrate solely on training the parameters of the graph projector. Noted that we also freeze the parameter of the graph encoder.

This progressive training approach ensures a coherent and incremental integration of diverse features.
At each stage, as the model parameters of previous stages are already optimized to incorporate existing modality features, 
we freeze its parameter in the next phase with the aim of integrating the new modality more effectively.
Extensive experiments also reveal the superiority of our training recipe over other alternatives.

At the inference stage of \modelname, 
for each instruction encompassing potential global contextual nodes and multiple target nodes to predict,
our framework integrates the outputs of the semantic embedding module and structural embedding module for each node into the LLM input,
and utilize the logit of \texttt{<label\_token>} after decoding as the final prediction score.

%% file: experiment.tex
\section{Experiments}
\label{sec:experiments}
In this section, we conduct comprehensive experiments to validate the effectiveness of the design choices in our framework.
\subsection{Experimental Setup}
\label{subsec:experimental setup}
\begin{table}[h!]
\caption{Dataset statistics.}
\centering
\small
\renewcommand{\arraystretch}{1.2} 
\setlength{\tabcolsep}{6pt}      
\begin{tabular}{p{1.2cm}| p{1.1cm}<{\centering} p{1.cm}<{\centering} p{1.4cm}<{\centering} p{1.7cm}<{\centering}}
\toprule
\textbf{Statistics}           & \textbf{WhoIsWho} & \textbf{MAG} & \textbf{TwiBot-20} & \textbf{SemEval-23F} \\ \midrule
\#Graphs                     & 1,664             & 2,316             & 1                    & 516                  \\
\#AvgNodes                     & 196              & 53               & 11,826                & 10                   \\
\%Anomaly                      & 11.79\%          & 12.43\%          & 55.72\%              & 28.42\%                \\
\#AvgLen                       & 83               & 45              & 631                  & 860                  \\
\#MaxLen                       & 10,774           & 276              & 2,190                  & 7,571\\
\#Train                        & 148,309          & 84,510            & 8,278                 & 361                  \\
\#Valid                        & 62,229           & -                & 2,365                 & 103                  \\
\#Test                         & 116,262          & 38,055            & 1,183                 & 52                   \\ 
\bottomrule
\end{tabular}
\label{tab:dataset_statistics}
\end{table}

\vpara{Datasets.}
We evaluate \modelname and baselines across four datasets from diverse domains: (1) \textbf{WhoIsWho}~\cite{zhang2024oag} , (2) \textbf{MAG}~\cite{roy2013microsoft}, (3) \textbf{TwiBot-20}~\cite{feng2021twibot}, and (4) \textbf{SemEval-23F}~\cite{piskorski2023semeval}. Both WhoIsWho and MAG originate from the task of incorrect assignment detection in author disambiguation. TwiBot-20 is a bot detection dataset built on the social graph of Twitter. SemEval-23F, on the other hand, focuses on identifying propaganda tactics used in ~\cite{DELL}.
The first three datasets are structured as binary classification tasks aimed at anomaly detection.
In contrast, SemEval-23F is a multi-label classification task targeting fine-grained identification of specific propaganda techniques. Detailed statistics for all datasets are provided in Table ~\ref{tab:dataset_statistics}.

\begin{table*}[t]
\caption{Overall results of anomaly detection across four datasets. Statistically significant improvements (p<0.05) over last stage are marked with *.}
    \centering
    \begin{tabular}{c|c|p{1.2cm}<{\centering}p{1.2cm}<{\centering}p{1.2cm}<{\centering}p{1.2cm}<{\centering}p{1.2cm}<{\centering}p{1.2cm}<{\centering}p{1.2cm}<{\centering}p{1.2cm}<{\centering}p{1.2cm}<{\centering}p{1.2cm}<{\centering}}

    \toprule
    Dataset & Metric & GCN & SOTA\newline~GNNs & RoBERTa & DeBERTa & Llama3-8B & Qwen2.5-7B & \modelname\newline-base & \modelname\newline+sem & \modelname\newline+graph \\
    \midrule
    \multirow{2}{*}{WhoIsWho}  
        & AUC & 0.655 & 0.741 & 0.649 & 0.654 & 0.741 & 0.751 & 0.757 & 0.771* & \textbf{0.789}* \\
        & MAP & 0.579 & 0.661 & 0.550 & 0.552 & 0.696 & 0.704 & 0.690 & 0.696* & \textbf{0.709}* \\
    \midrule
    \multirow{2}{*}{MAG}       
        & AUC & 0.670 & 0.839 & 0.899 & 0.873 & 0.958 & 0.960   & 0.949 & 0.961* & \textbf{0.963}* \\
        & MAP & 0.545 & 0.765 & 0.831 & 0.795 & 0.923 & 0.924   & 0.910 & \textbf{0.932}* & 0.931 \\
    \midrule
    \multirow{2}{*}{TwiBot-20}  
        & AUC & 0.847 & 0.932 & 0.925 & 0.907 & 0.943 & 0.944 & 0.916 & 0.943* & \textbf{0.945}* \\
        & MAP & 0.833 & 0.928 & 0.927 & 0.918 & 0.942 & 0.943  & 0.912 & 0.942* & \textbf{0.945}* \\
    \midrule
    \multirow{2}{*}{SemEval-23F}
        & AUC & 0.815 & 0.815 & 0.805 & 0.843 & 0.874 & \textbf{0.875} & 0.867 & \textbf{0.875}* & \textbf{0.875} \\
        & MAP & 0.748 & 0.748 & 0.740 & 0.784 & 0.809 & 0.792 & 0.796 & 0.811* & \textbf{0.815}* \\
    \bottomrule
    \end{tabular}
    \label{tab:main_result}
\end{table*}
\vpara{Baselines.}
We conduct a comparative analysis of our model against the popular GNN-based methods and LM-based methods:
\begin{itemize}[leftmargin=*]
\item {\textbf{GCN}} ~\cite{gcn}:
A two-layer GCN model is used to learn node features, followed by a classifier to categorize the nodes.
\item {\textbf{SOTA-GNNs}} ~\cite{gccad,DBLP:conf/www/HuDWS20}:
We use GCCAD~\cite{gccad} (node-edge contrastive learning for anomaly detection) for WhoIsWho and MAG datasets, and HGT~\cite{DBLP:conf/www/HuDWS20} (heterogeneous social network representation learning) for the TwiBot-20 dataset.
We further conduct additional experiments to find the best-performing GNNs, with results detailed in the Appendix ~\ref{sec:sota-gnns}.
\item {\textbf{RoBERTa}} ~\cite{roberta}:
We finetune a RoBERTa-base model 
followed by 
a trainable head for node classification. 
\item {\textbf{DeBERTa}~\cite{deberta}: We further leverage a pretrained DeBERTa-v3-large model for sequence classification.
Given the challenges of fully fine-tuning the DeBERTa model on long contexts, we apply LoRA~\cite{lora} fine-tuning for  the DeBERTa model.}
\item {\textbf{Llama3-8B~\cite{meta2024introducing} and Qwen2.5-7B~\cite{DBLP:journals/corr/abs-2412-15115}}}: We adopt widely used open-sourced LLM models, Llama3-8B and Qwen2.5-7B. These models directly take all textual features as  inputs, and use the generated logits of each \texttt{<label\_token>} to obtain the final prediction results.

\end{itemize}

Our model has three variants, corresponding to the three sequential training phases:
(1) \textbf{\modelname-base} represents the base version of the model, where only key textual features are used as LLM input, as described in Section~\ref{subsec:base};
(2) \textbf{\modelnamesem} indicates the integration of the semantic embedding module on top of the base model, as described in Section~\ref{subsec:textemb};
(3) \textbf{\modelnamegraph} represents the addition of the structural embedding module after the semantic embedding module has been incorporated, as described in Section~\ref{subsec:graphemb}.

\vpara{Evaluation Metrics.}
We adopt AUC and MAP as the evaluation metrics, which are widely used in related anomaly detection tasks~\cite{whoiswho, gccad}.
For the IND task (i.e., WhoIsWho and MAG), we follow the approach in~\cite{zhang2024oag} to compute the final results by weighting the outliers across all authors. Specifically, the weight for each author is determined by the proportion of their incorrect instances relative to the total number of incorrect instances.
For the multi-label classification task (i.e., SemEval-23F), the overall score is calculated by weighting each category based on the number of anomalies in the specific category.

\subsection{Implementation Details}
\label{sec:details}
As for our framework, we utilize the Llama3-8B~\cite{touvron2023llama,meta2024introducing} to train a base model.
For the semantic embedding module, we utilize the full information of each sample as raw input text, and feed it into small language models (RoBERTa or DeBERTa) to get the summarized input embeddings.
For the structural embedding module, we utilize the SOTA GNNs in each task as graph encoder to obtain the structural embedding.

For the WhoIsWho, MAG, TwiBot, and SemEval-23F datasets, the number of turns is set to 10, 10, 8, and 6, respectively. We select the key attribute based on experimental observations or knowledge derived from existing works. For the WhoIsWho and MAG datasets, we conduct experiments on different features (as shown in Figure \ref{fig:feature} and Figure \ref{fig:mag_feature}), to identify the key features for the corresponding datasets. We use Title and Author as the key attributes for the WhoIsWho dataset. For the TwiBot-20 dataset, inspired by \cite{DBLP:conf/wsdm/CaiT0ZWZL24}, we use metadata as the key feature. For the SemEval-23F dataset, we select the first 512 tokens of the full text as the key attribute.
Further implementation details are shown in Appendix ~\ref{sec:details_appendix}.

\hide{
stage3 compare with sota methods 
For WhoIsWho datasets
AUC t-statistic: -, p-value: -
MAP t-statistic: -, p-value: -
For TwiBot datasets
AUC t-statistic: 2.390457218668787, p-value: 0.04382455063702764
MAP t-statistic: 2.286190426597742, p-value: 0.05157060544466868
}
\subsection{Main Results}
\label{sec:main_result}

Table \ref{tab:main_result} provides a holistic comparison of different anomaly detection methods on four datasets. 
Generally speaking, LLM-based methods (Llama3-8B, Qwen2.5-7B, and our model \modelname) outperform small language model (SLM)-based methods and GNN-based methods.
GNN-based methods are competitive on WhoIsWho and TwiBot-20 datasets due to the importance of structural relationships like coauthor links and social connections, while semantic features play a more crucial role in MAG and SemEval-23F datasets.
We observe that SLM-based methods consistently underperform LLM-based methods on all datasets, demonstrating the remarkable fitting and generalization ability of LLMs.

Although fine-tuned LLMs yield satisfactory performance, they face challenges including limited context utilization and high computational costs.
\modelname-base achieves similar results to Llama3-8B using only key textual features and multi-turn instruction templates for improved training and inference efficiency.
Further integration of the semantic embedding module (\modelname-sem) achieves state-of-the-art performance while being more compute-efficient than traditional LLM fine-tuning approaches.
Ultimately, incorporating the structural embedding module further enhances the model's efficacy by addressing LLMs' limitations in capturing graph structural features.
The main results validate the effectiveness of our framework, and we conduct extensive ablation studies to verify the rationality of our design choices.

\subsection{Ablation Studies}
\label{subsec:ablation study}

\vpara{Effect of different foundation models.}
We examine the effect of different backbone LLMs in our framework, including 
Qwen2.5-7B~\cite{qwen2} and Llama3-8B~\cite{touvron2023llama}.
As shown in Table \ref{tab:foundation model},
For all backbone models in the WhoIsWho dataset, a consistent stepwise outperformance is observed in each stage, 
demonstrating the effectiveness of our three-stage training strategy, 
which robustly integrates multi-source features in a progressive manner. 
In the TwiBot-20 dataset, the integration of graph features yields only minor improvements, possibly because the impact of textual features overshadows the graph features.
\begin{table}[h!]
    \caption{Ablation studies on different foundation models in WhoIsWho and TwiBot-20 datasets.
    }
    \begin{tabular}{c|c|cc|cc}%
    \toprule
    \multicolumn{2}{c|}{\textbf{Dataset}} & \multicolumn{2}{c|}{\textbf{WhoIsWho}} & \multicolumn{2}{c}{\textbf{TwiBot-20}} \\
    \midrule
    Model & \component & AUC & MAP & AUC & MAP \\
    \midrule
    \multirow{3}{*}{Qwen2.5-7B} & base & 0.724 & 0.664 & 0.907 & 0.903 \\
    & +sem & 0.773 & 0.696 & \textbf{0.947} & \textbf{0.950} \\   
    & +graph & 0.780 & 0.702& 0.946 & 0.949 \\
    \midrule 
    \multirow{3}{*}{Llama3-8B} & base & 0.757 & 0.690 & 0.916 & 0.912 \\
    & +sem & 0.771 & 0.696 & 0.943 & 0.942 \\    
    & +graph & \textbf{0.789} & \textbf{0.709} & 0.945 & 0.945 \\
    \bottomrule
    \end{tabular}
    \label{tab:foundation model}
\end{table}

\begin{figure}[h!]
    \centering
    \includegraphics[width=1.0\linewidth]{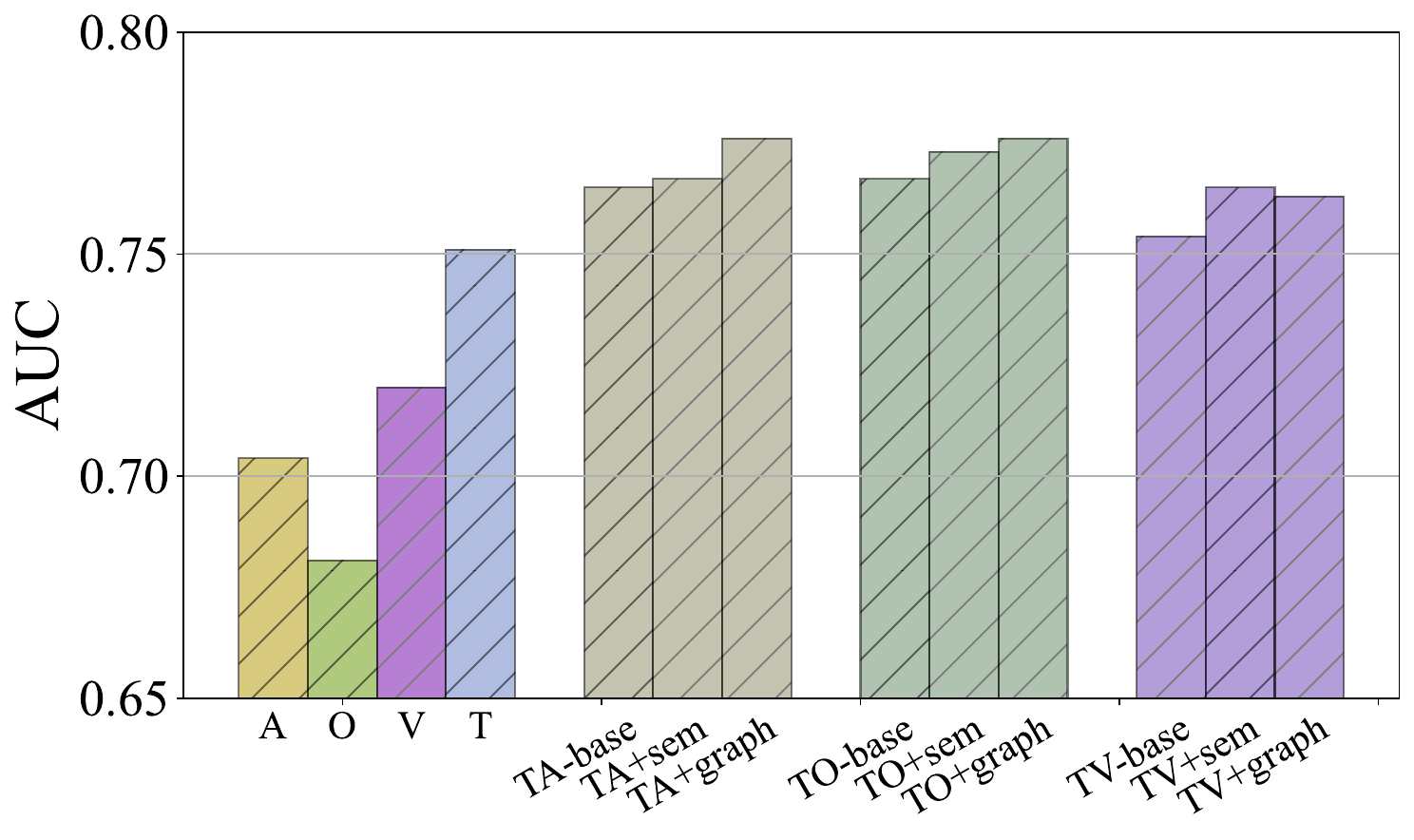}
    \caption{Ablation studies on different paper attributes as the LLM input for WhoIsWho dataset.
    \textnormal{``A'', ``O'', ``V'', and ``T'' denote using the author, organization, venue, or title attribute from a paper as key information input into the LLM, respectively. Similarly, ``TA'', ``TO'', and ``TV'' represent combinations of two of these features as the key input.}}
    \label{fig:feature}
\end{figure}

\vpara{Effect of different paper attributes.}
Figure \ref{fig:feature} illustrates the impact of 
paper attributes used in our framework in the WhoIsWho dataset.
We first alter the input paper attributes in the first stage of our framework (without semantic embedding and structural embedding module).
We observe that 
``paper title'' is the best-performing attribute,
followed by paper venues.
The reason may lie in that titles and venues encompass rich domain-specific information, which can be well-captured by the LLM.
In contrast, organization and author attributes are ad-hoc personal features, 
often used as effective co-occurred structural features,
which are arduous to be utilized by the LLMs.
Furthermore, the superiority of the title attribute over the venue is also partly attributed to the finer-grained topical information conveyed in titles.
To further investigate the synergistic effects between attributes, 
we conduct the complete training stages by employing the best single attribute (i.e., title) and another paper attribute.
``Title+venue''  has a small edge over the single usage of the paper title ($+0.3\%$ in terms of AUC),
possibly due to that titles and venues embody similar domain-specific features.
By contrast, both ``Title+org'' and ``Title+author'' each deliver significant improvements over the single title attribute, showing $+1.6\%$ and $+1.4\%$ increase in terms of AUC, respectively. These improvements are likely due to the complementary nature of the title attribute when combined with either ``org'' or ``author''.

We also conducted in-depth experimental analyses of various features on the MAG dataset, as referenced in Appendix ~\ref{sec:mag_feature}.

\begin{figure}[]
    \centering
    \includegraphics[width=0.35\textwidth]{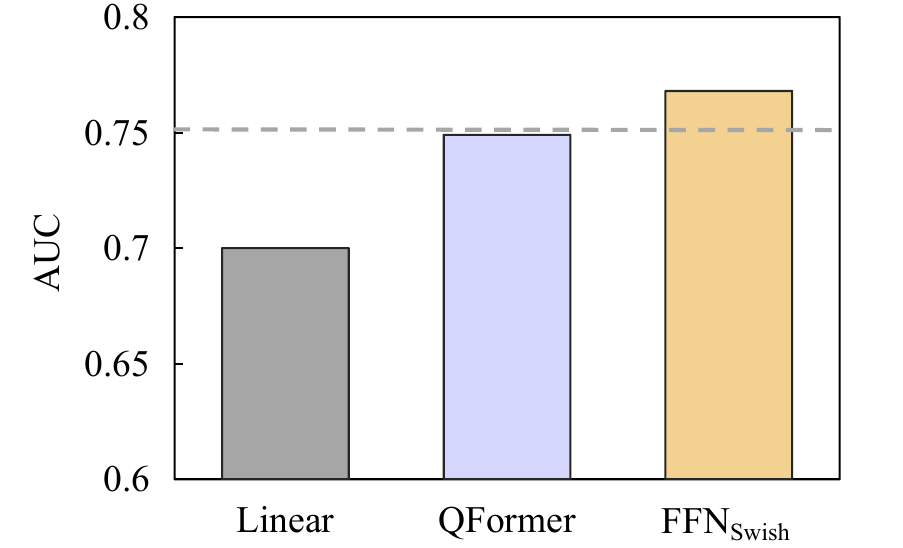} 
     \caption{Ablation studies on the different text projectors on the WhoIsWho dataset. \textnormal{The gray dashed line represents the performance of \modelname-base, which serves as the foundation for training the text predictors.}}
    \label{fig:projector}
\end{figure}

\vpara{Effect of different text projectors.}
\label{sec:projector}
We try several types of text projectors with increased model complexity,
including a single linear layer, a two-layer feed-forward network $\text{FFN}_{\text{Swish}}$, and Q-Former~\cite{blip2}. 
The Q-Former is frequently used in multi-modal LLMs to align multimodal information (such as text and vision).
Here Q-Former is parameterized by a single-layer Transformer decoder with a cross-attention structure.
The experimental results on WhoIsWho are presented in Figure \ref{fig:projector}. 
Among the three projectors, only $\text{FFN}_{\text{Swish}}$ exhibits noticeable improvements over \modelname-base. 
Linear projector impairs performance owing to its lack of expressive power.
Unexpectedly, Q-Former is inferior to $\text{FFN}_{\text{Swish}}$.
The reason might be that the Q-former has an excessive number of parameters as a projector, making it difficult to train.

\begin{table}[h]
    \centering
    \caption{Influence of the number of turns in the multi-turn instruction template on two datasets. \textnormal{The best values for training and inference time (min), and AUC are highlighted in \textbf{bold}.}}
    \begin{tabular}{c|ccc|ccc}
        \toprule
        \multirow{2}{*}{$\#turns$} & \multicolumn{3}{c|}{\textbf{WhoIsWho}} & \multicolumn{3}{c}{\textbf{TwiBot-20}} \\
        \cmidrule{2-7}
        & \textbf{Train} & \textbf{Inf} & \textbf{AUC} & \textbf{Train} & \textbf{Inf} & \textbf{AUC} \\
        \midrule
        1  & 315.0 & 56.4 & 0.744          & 3.32  & 0.21 & 0.911          \\
        2  & 169.8 & 28.8 & 0.744          & 1.76  & 0.11 & \textbf{0.916} \\
        4  & 91.8 & 15.0 & 0.751          & 1.13  & 0.08 & 0.912          \\
        8  & 49.8 & 7.8 & \textbf{0.763} & 0.97  & 0.07 & 0.913          \\
        16 & \textbf{25.2} & \textbf{4.2} & 0.737  & \textbf{0.93} & \textbf{0.07} & 0.915 \\
        \bottomrule
    \end{tabular}
    \label{tab:multi_turn}
\end{table}

\vpara{Effect of $\#turns$ in each instruction.}
We further study how many target nodes to predict in each multi-turn instruction ($\#turns$: i.e., the number of turns) is the best in terms of model accuracy and efficiency.
In this part, we take the WhoIsWho and TwiBot-20 datasets as examples to analyze the scenarios with and without the global context, respectively.
Table \ref{tab:multi_turn} shows the accuracy (i.e., AUC) and efficiency performance with $\#turns$ in the range of $\{1, 2, 4, 8, 16\}$.
We observe that an increasing number of multi-turns inversely correlates with training and inference time, leading to improved speed as the number of turns grows.
This is because the multi-turn mechanism 
enhances the reuse efficiency of global contextual information
by sharing it across turns, thereby resulting in speed gains. 
Additionally, it can be observed that the AUC metric of the model reaches its optimal value when \#turn is set to 8 in the WhoIsWho dataset. 
The accuracy improvement of multi-turns over single-turn validates the effectiveness of the ``soft'' demonstration mechanism in providing additional semantic demonstrations for decoding subsequent papers.

For the TwiBot-20 dataset,
even in the absence of shared global context, notable training and inference speed improvements are still observed. 
This indicates that the multi-turn strategy remains effective in such scenarios. These speed gains are similar to those achieved by an efficient sequence packing strategy~\cite{krell2021efficient,DBLP:conf/emnlp/BaiLZHQH0DL24}.

\vpara{Effect of three-stage training strategy}
We compare our progressive training strategy with several alternative strategies to inject multi-modal information.
(1) \textbf{Reverse} of two-modal training:
injecting the structural embedding module first followed by the
semantic embedding module;
(2) \textbf{From-scratch} training of all projectors:
injecting semantic embedding module and structural embedding module simultaneously and training text projector and graph projector jointly.
\begin{figure}[h]
\centering
\includegraphics[width=\linewidth]{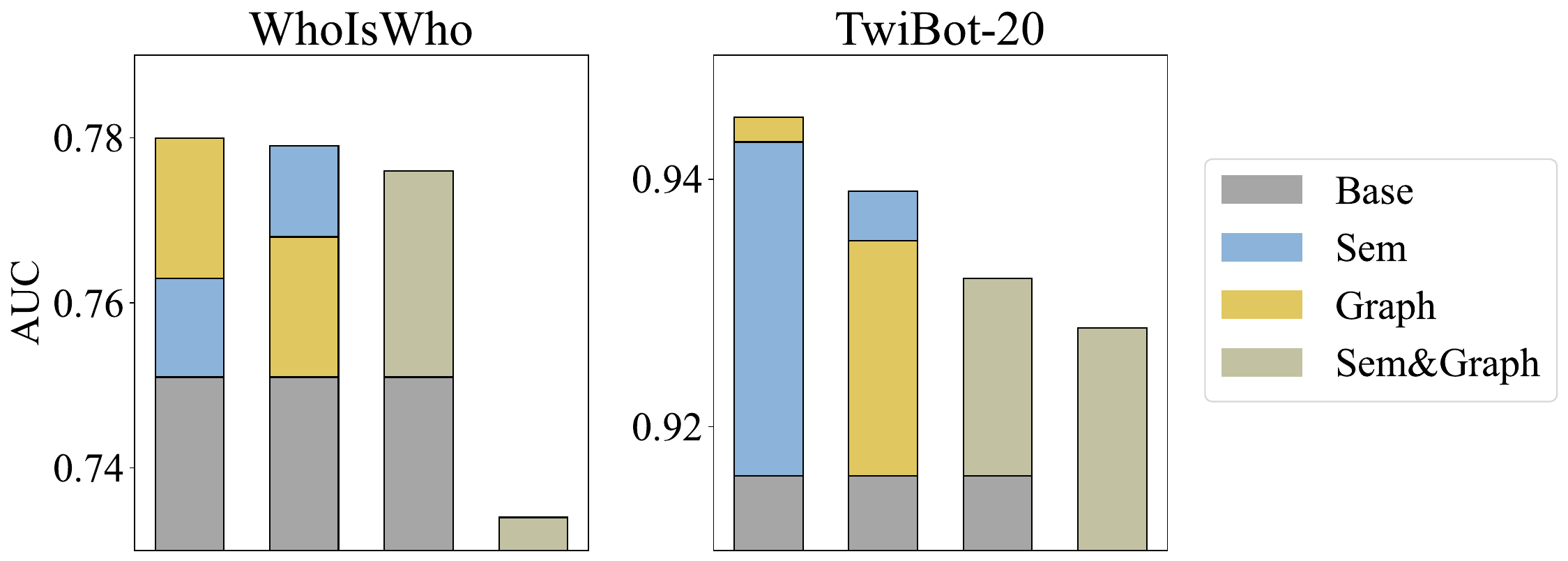} 
\caption{The AUC comparison of different training strategies on two datasets: WhoIsWho (left) and TwiBot-20 (right).}
\label{fig:strategies}
\end{figure}

As shown in Figure \ref{fig:strategies}, the experimental results indicate that concurrently training the text projector and graph projector yields suboptimal improvement over \modelname-base, 
but underperforms both progressive and reverse strategies.
We speculate that it is difficult to achieve optimal parameters for the two projectors simultaneously in from-scratch training.
In addition, reserve of two-modal training yields stable performance improvements over \modelname-base, but it is slightly less effective than first adding the semantic embedding module followed by the structural module.
\begin{figure}[h!]
    \centering
    \begin{subfigure}[b]{0.225\textwidth} 
        \centering
        \includegraphics[width=\textwidth]{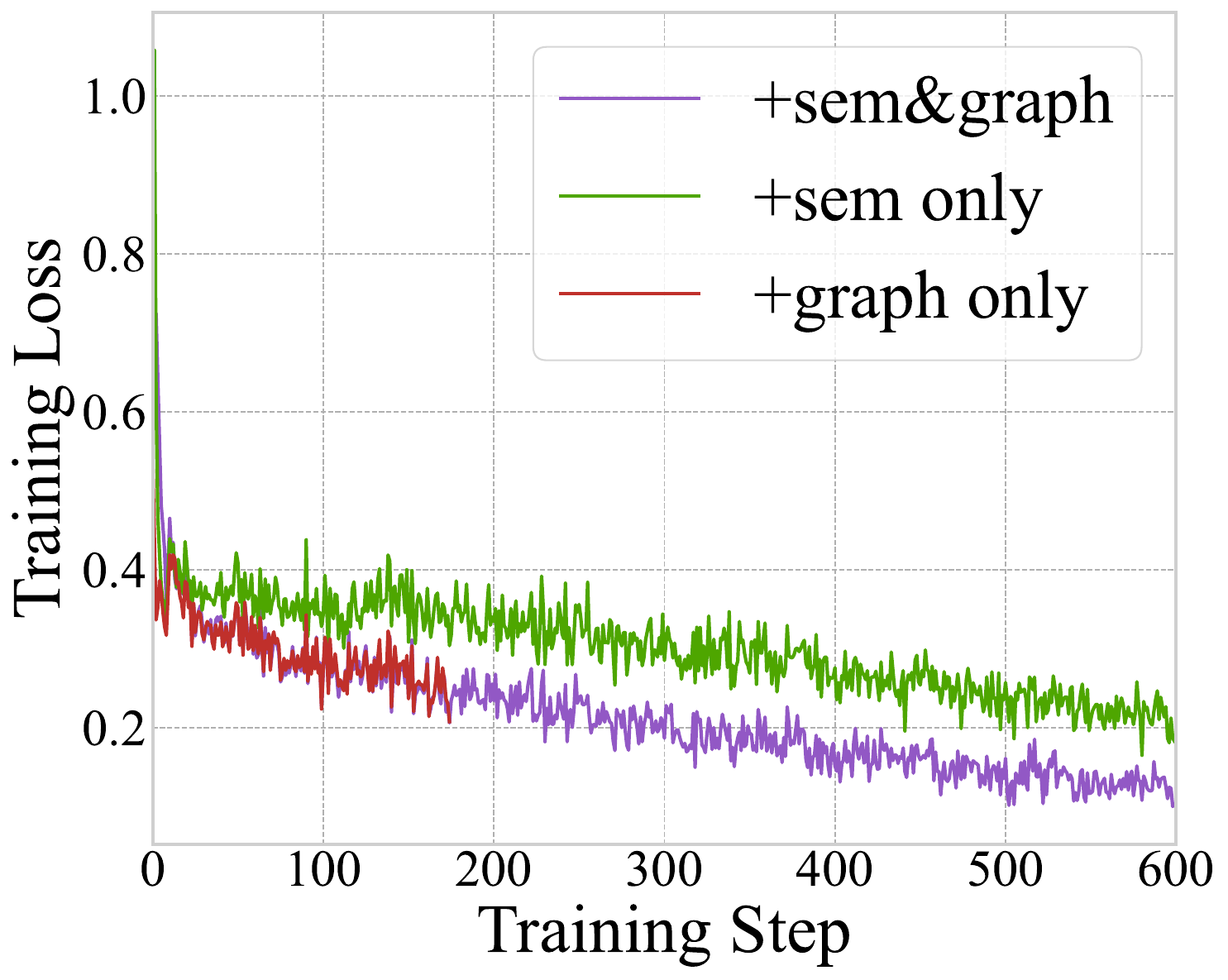} 
        \caption{Training Loss}
        \label{fig:subfig1} 
    \end{subfigure}
    \hfill 
    \begin{subfigure}[b]{0.225\textwidth} 
        \centering
        \includegraphics[width=\textwidth]{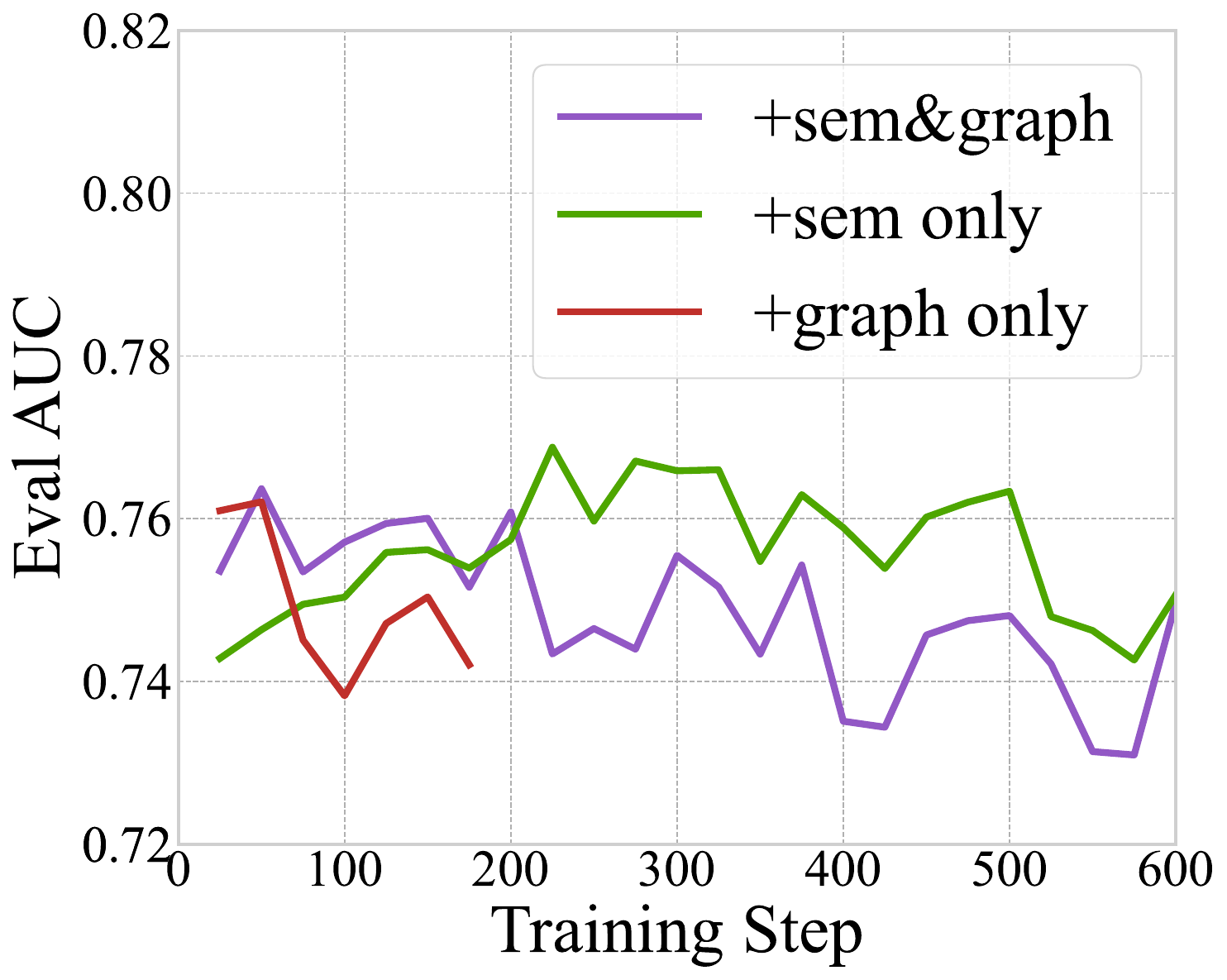} 
        \caption{AUC on validation set}
        \label{fig:subfig2} 
    \end{subfigure}
     \caption{Comparison between individually trained models and joint training models in terms of training loss and AUC.
     }
    \label{fig:ETandTS}
\end{figure}

In Figure \ref{fig:ETandTS}, we further visualize the changes in training loss and AUC on the validation set during the training process
concerning from-scratch training and two-modal training (with semantic or structural embedding modules) on the WhoIsWho dataset.
For two-modal training,
the text projector converges at about $250$ training steps
while the graph projector converges at only about $50$ training steps.
However, from-scratch training, in terms of both AUC and training loss, appears to have learned predominantly from a single source of graph features.

\subsection{Model Efficiency}
\begin{table}[]
    \caption{
    The efficiency performance of our model variants and 
    the full-text fine-tuned counterparts
    (in minutes).
    }
    \centering
    \resizebox{\columnwidth}{!}{
    \begin{tabular}{c|cc|cc}
        \toprule
        \multirow{2}{*}{Time Cost} & \multicolumn{2}{c|}{WhoIsWho} & \multicolumn{2}{c}{TwiBot-20} \\
        \cmidrule(lr){2-3} \cmidrule(lr){4-5}
         & \makecell{Train } & \makecell{Inference} & \makecell{Train} & \makecell{Inference} \\
        \midrule
        \modelname-base & 29.40 & 6.00 & 0.90 & 0.07 \\
        +sem (stage 2) & 37.20 & 10.80 & 0.90 & 0.07 \\
        +graph (stage 3) & 39.00 & 12.00 & 0.91 & 0.07 \\
        \midrule
        Llama3-8B & 592.80 & 130.20 & 3.11 & 0.21 \\
        \bottomrule
    \end{tabular}
    }
    \label{tab:efficiency}
\end{table}
\label{subsec:model_efficiency}
In this subsection, 
we compare the time efficiency of the \modelname model with LLM-based approaches.
All the experiments utilize the Llama3-8B model.
Table \ref{tab:efficiency} presents the detailed time costs for training and inference. 
The training and inference time are defined as the time required to perform a single full pass of training (or inference) over the training and validation datasets, respectively.
Compared to vanilla Llama, our base model leverages a shorter input length as well as multi-turn instructions,
achieving both fast training and inference speed on both the WhoIsWho and TwiBot-20 datasets. Notably, on the WhoIsWho dataset, our model achieves over a 10$\times$ speedup in inference and a 5$\times$ speedup in training.
Even without shared global context, our method still achieves a 3$\times$ speedup in inference on the TwiBot-20 dataset.

\subsection{Model Ensemble}
\label{ensemble}

\begin{figure}[h]
    \centering
    \includegraphics[width=7cm]{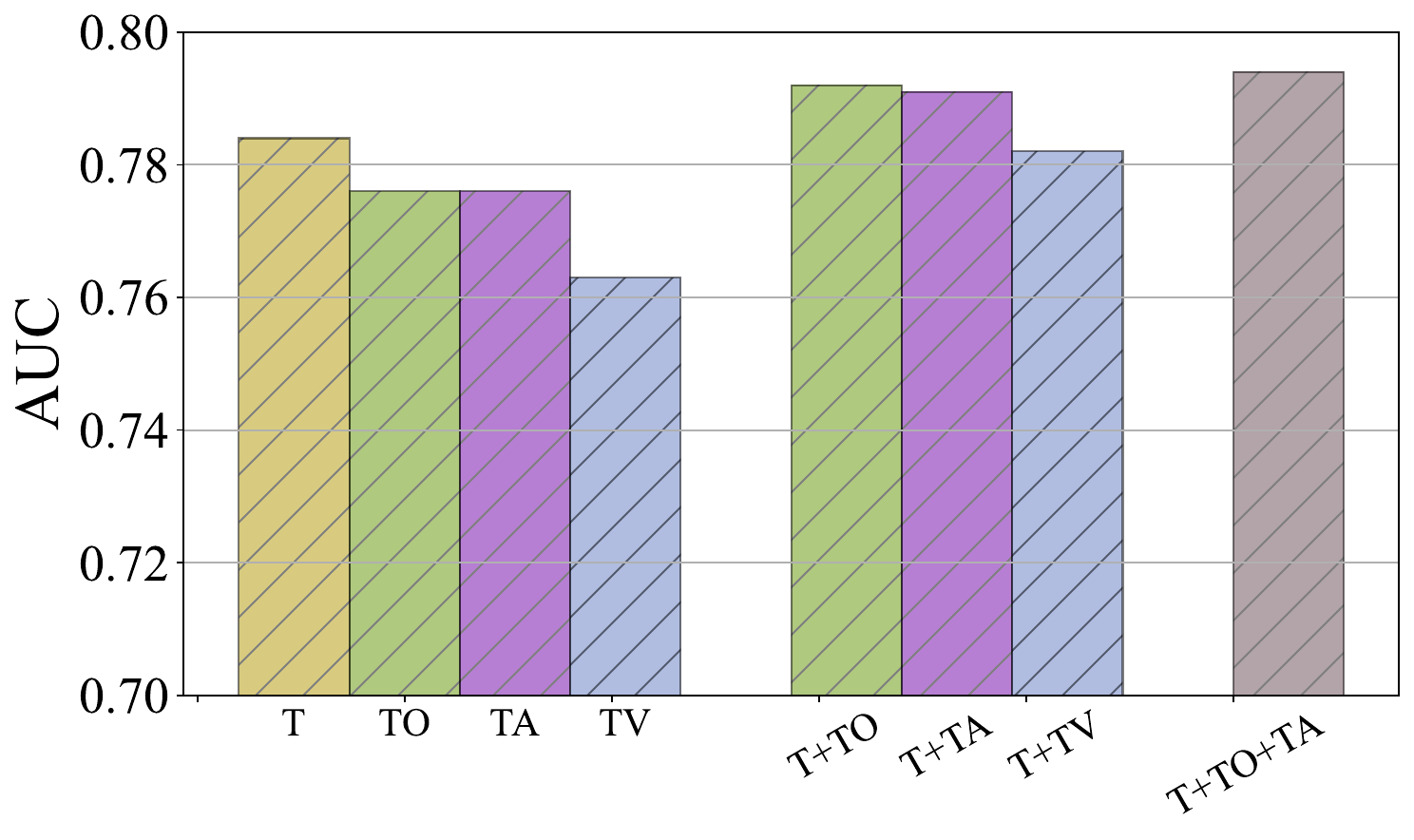}
    \caption{Performance comparison of different ensemble models by utilizing various paper attributes. \textnormal{ }}
    \label{fig:ensemble}
\end{figure}

To unlock the potential of our framework, we further perform model ensemble by using different paper attributes. As shown in Figure 8, the best two-model ensemble results are obtained by combining the title attribute (T) with the title and organization attributes (T+TO), resulting in a $0.8\%$ AUC improvement compared to the single title attribute baseline. The combination of title and author attributes (T+TA) also yields decent results with a $0.7\%$ AUC increase, indicating that incorporating raw organization texts into the LLM inputs exhibits preferable performance compared with injecting these information into the semantic embedding module. 

Ultimately, by integrating the best three models (T+TO+TA), we achieve the optimal performance. We further compare our method with the top three winning solutions from KDD Cup 2024. Our approach demonstrates advantages in both performance and efficiency compared to LLM-based methods, while the GCN+LGBM solution shows speed advantages due to smaller model scale but requires substantial feature engineering effort.

\begin{table}[]
\small
    \centering
    \caption{Performance comparison of models using different categories of features: Top-3 on the Whoiswho-IND test leaderboard in KDD Cup 2024.}
    \begin{tabular}{p{0.6cm}<{\centering}|c|p{0.7cm}<{\centering}p{0.6cm}<{\centering}p{0.7cm}<{\centering}|p{0.7cm}<{\centering}c}
    \toprule
         Rank&Model&\makecell{Man-\\ual}&\makecell{Text-\\ual}&\makecell{Struc-\\tural}& \makecell{AUC\\(\%)}&\makecell{Test\\Time (h)} \\
         \midrule
         1~\cite{rank1}& GLM3~\cite{glm130,glms}& -  &\checkmark & - & 83.45 &>10\\
         \midrule
         \multirow{2}{*}{2~\cite{rank2}}&GCN~\cite{gcn}& \checkmark & \checkmark & \checkmark&\multirow{3}{*}{82.49}& \multirow{3}{*}{<0.1}\\ 
         & LGBM~\cite{lgb}&\checkmark & \checkmark  &- & & \\
         \midrule
         \multirow{4}{*}{3~\cite{rank3}}&GLM3~\cite{glm130,glms}& -  &\checkmark & - & \multirow{4}{*}{81.35}&\multirow{4}{*}{>10} \\
         & GLM4~\cite{glm130,glms}& - & \checkmark & - &&\\
         & Mistral~\cite{mistral}& - & \checkmark & - && \\
         & LGBM~\cite{lgb}&\checkmark & \checkmark  &- && \\
         \midrule
         -&\modelname &  -  &\checkmark&\checkmark & \textbf{83.51}&1.3\\
         \bottomrule
    \end{tabular}
    \label{tab:leaderboard}
\end{table}

%% file: conclusion.tex
\section{Conclusion}
The proposed text-rich and graph-informed language model \modelname effectively addresses the anomaly detection problem by integrating both structural and semantic features. 
\modelname is a multi-modal-like multi-turn instruction tuning framework, 
which includes task-guided multi-turn instruction tuning, 
semantic embedding module with rich attributes, 
and a structural embedding module.
Extensive experiments reveal that our model outperforms 
previous state-of-the-art graph-based methods and fine-tuned language methods,
and demonstrate significant training efficiency and inference efficiency compared with previous LLM-based methods.
Our idea of task-guided instruction tuning on the 
text-rich and graph-informed
language model exhibits the potential to foster future research toward addressing downstream tasks of text-rich graphs.

%% file: appendix.tex
\section{Implementation Details}
\label{sec:details_appendix}

During the training of the LLaMA3-8B and Qwen 2.5-7B models, we set the maximum input length to 8K, and use FlashAttention~\cite{dao2022flashattention,dao2023flashattention} to reduce the memory consumption during training and inference. We consistently adopted LoRA~\cite{lora} with a rank of 8, LoRA alpha value of 16, and a dropout rate of 0.05. On the WhoIsWho dataset, the per-device batch size was set to 1, with a gradient accumulation step of 16, resulting in a global batch size of 128 for 8-card training. All experiments were conducted using Nvidia A100 GPUs, each with 80GB of memory. Except for the SemEval-23F dataset, which, due to its relatively small size, was trained using 2 GPUs, all other experiments utilized the full set of GPUs available.

We used AdamW~\cite{adamw} as the optimizer with a weight decay of 1e-3. A cosine learning rate scheduler with a linear warm-up was employed for the \modelname-base model, with a warm-up ratio of 0.1 and a peak learning rate of 1e-4. In contrast, a constant learning rate of 5e-5 was applied for the “+sem” and “+graph” models. The “base”, “+sem”, and “+graph” models were trained for 6, 10, and 4 epochs, respectively, with evaluations conducted every 25 global steps throughout the training process. Each phase of the training procedure leveraged the best-performing parameters from the evaluation of the prior phase. For the FFN layer in the semantic/structural embedding module, the intermediate hidden size was set to twice the hidden size of the following LLM.





\section{Effect of different feature on MAG dataset}
\label{sec:mag_feature}
\begin{figure}[h!]
    \centering
    \includegraphics[width=1.0\linewidth]{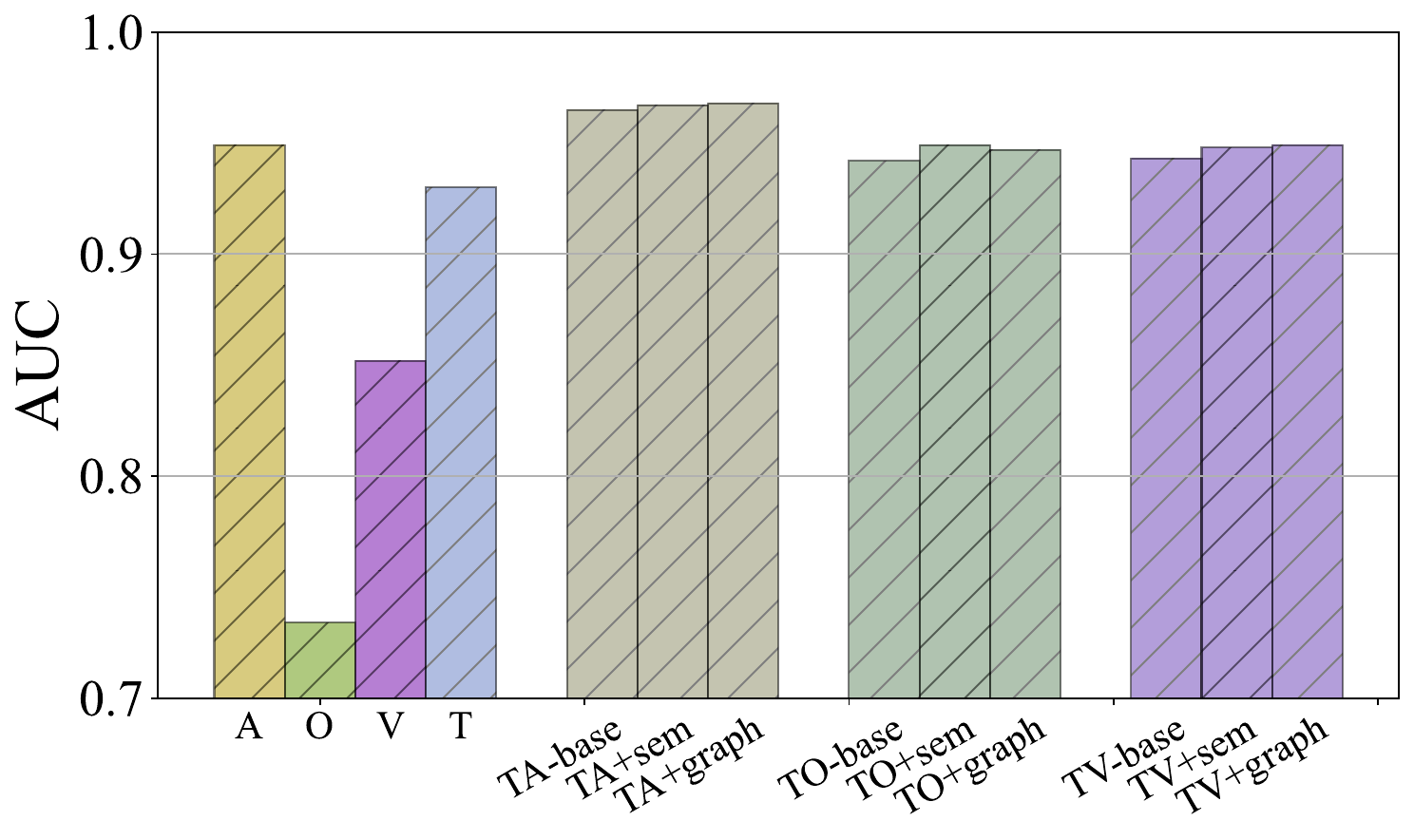}
    \caption{Ablation studies on different paper attributes as the LLM input for MAG dataset.
    \textnormal{``A'', ``O'', ``V'', and ``T'' denote using the author, organization, venue, or title attributes from a paper as key information input into the LLM, respectively. Similarly, ``TA'', ``TO'', and ``TV'' represent combinations of two of these features as the key input.}}
    \label{fig:mag_feature}
\end{figure}
In contrast, the MAG dataset exhibits a distinct pattern where author information performs best, followed by titles, while organizational affiliations show the weakest results. 
This discrepancy may arise from two factors: the MAG dataset's partial absence of certain metadata features, and the inherent challenges for LLMs to effectively utilize structural features like author co-occurrence patterns and organizational relationships. Across both datasets, titles consistently outperform venues, likely due to their ability to convey finer-grained topical information through specialized terminology and precise phrasing.
For the combination of different feature,``Title+author'' achieves the best base model, this is because both the author and title features demonstrate strong performance on their own, contributing significantly to the model's predictive capabilities. On the other hand, even though the organization (org) and venue features individually do not perform particularly well, incorporating them into the model alongside the title feature still brings marginal improvements. These slight gains indicate that org and venue provide complementary information that helps the model capture additional nuances in the data, leading to an overall enhancement in performance.

\begin{table}[h]
    \caption{
    The efficiency performance of our model variants and the best baseline
    (in hours).}
    \centering
    \resizebox{\columnwidth}{!}{
    \begin{tabular}{c|ccc}
        \toprule
        Method
        & \makecell{Training \\time/epoch} & \makecell{Convergence\\time} & \makecell{Inference\\time}\\
        \midrule
        \modelname-base & 0.49& 0.81 & 0.10  \\
        +sem (stage 2)& 0.62& 1.82& 0.18 \\
        +graph (stage 3) & 0.65& 0.63 &0.20 \\
        Total & \textbf{0.65}& \textbf{3.25} & \textbf{0.20} \\
        \midrule
        \rankonename & 17.76\hide{(non balanced)} &  8.88 & 1.95\\
        \bottomrule
    \end{tabular}
    }
    \label{tab:efficiency_whoiswho}
\end{table}

\section{Experiment on Scaling Law}
We further explored the scaling law of the Qwen2.5 model on two representative datasets (3B, 7B, 14B). For all three models, we adopted the same LoRA configuration with a rank of 8 and an alpha of 16, and keeping other hyperparameters consistent. The experimental results are shown in the table below. We observed that the 7B model performed the best on both tasks, while the 14B model only slightly outperformed the 3B model.

\begin{figure}[h]
\centering
\begin{subfigure}[t]{0.47\columnwidth}
    \centering
    \includegraphics[width=\linewidth]{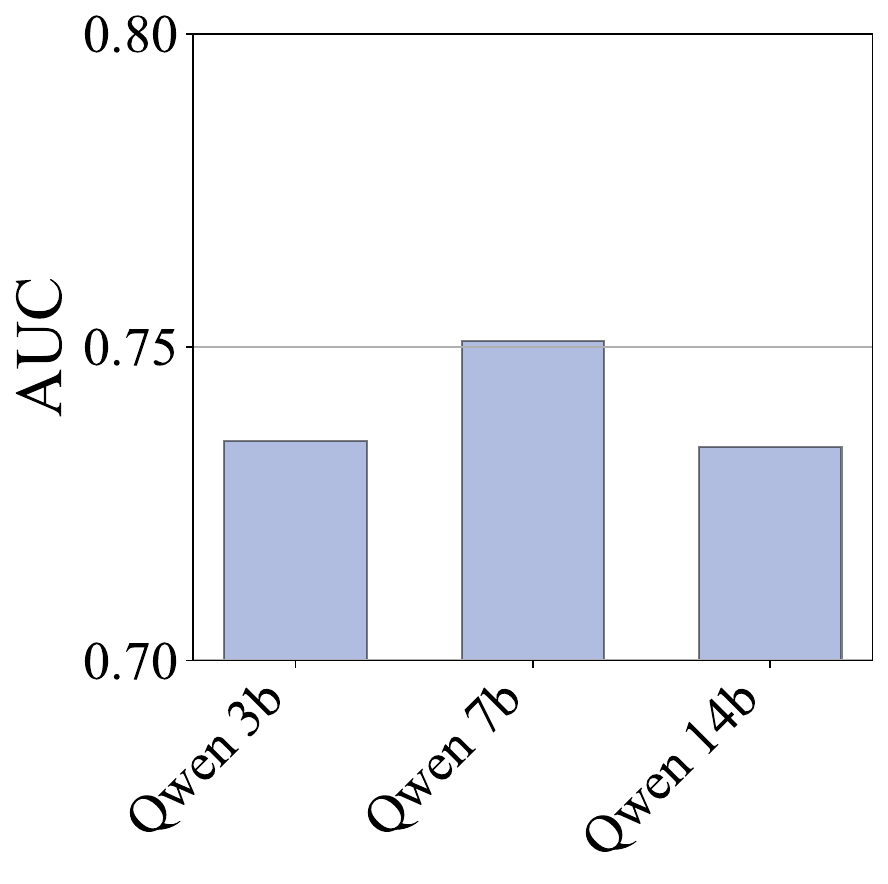}
    \caption{WhoIsWho}
    \label{fig:whoiswho_scaling}
\end{subfigure}
\hfill
\begin{subfigure}[t]{0.47\columnwidth}
    \centering
    \includegraphics[width=\linewidth]{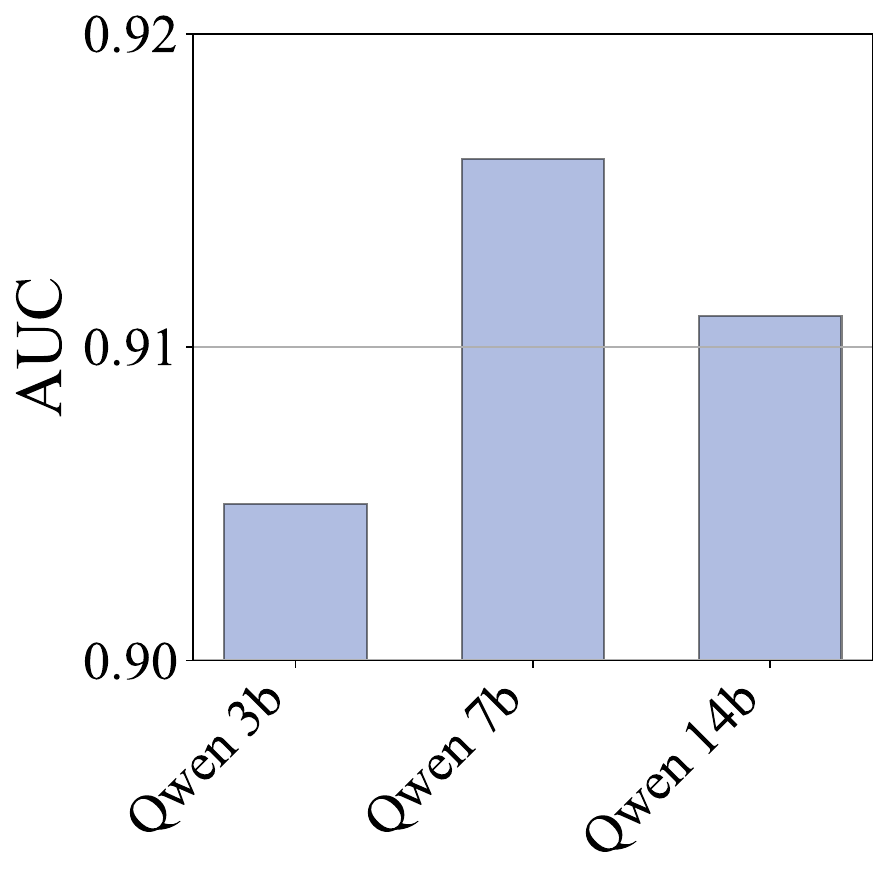}
    \caption{TwiBot-20}
    \label{fig:bot_scaling}
\end{subfigure}
\caption{Comparison of different model scales (3B, 7B, 14B) on two datasets: WhoIsWho (left) and TwiBot-20 (right).}
\label{fig:scaling_law}
\end{figure}

\section{Selection of SOTA-GNN Models}
\label{sec:sota-gnns}
We further compared various GNN baselines~\cite{DBLP:conf/icml/JuMYQG000024,DBLP:conf/iclr/ZhuoLHHTFC24} and language model~\cite{roberta,li2023towards} as node feature encoder on each dataset and adopted the best-performing model as the SOTA-GNNs for each section. We present the models we compared in Table \ref{tab:sotagnns}, and evaluated the embeddings from both pre-trained models and large language models. Among them, the GCCAD and HGT models were meticulously fine-tuned on each dataset, thus yielding robust results.

\begin{table}[h]
\centering
\caption{Performance comparison of models with different embeddings on WhoIsWho and TwiBot20 datasets.}
\label{tab:sotagnns}
\begin{tabular}{lc|cc|cc}
\hline
\multirow{2}{*}{Model} & \multirow{2}{*}{Embedding} & \multicolumn{2}{c}{WhoIsWho} & \multicolumn{2}{c}{TwiBot20} \\
 &  & AUC & MAP & AUC & MAP \\ \hline
HEAL          & RoBERTa  & 0.728 & 0.639 & 0.791 & 0.754 \\
PMP           & RoBERTa  & 0.743 & 0.674 & 0.920 & 0.914 \\
GCCAD         & RoBERTa  & 0.741 & 0.661 & -     & -     \\
GCCAD         & GTE      & 0.716 & 0.643 & -     & -     \\
HGT           & RoBERTa  & -     & -     & 0.932 & 0.928 \\
HGT           & GTE      & -     & -     & 0.937 & 0.932 \\ \hline
\end{tabular}
\end{table}
